\documentclass[a4paper,oneside,12pt,english]{article}

\usepackage[pdftex]{color,graphicx}

\usepackage[T1]{fontenc}
\usepackage[latin9]{inputenc}
\usepackage{url}

\usepackage{amsmath}
\usepackage{amssymb}

\makeatletter
\usepackage[backref=page]{hyperref}
\hypersetup{plainpages = false,
              breaklinks=true,
              a4paper=true,
              linktocpage,
              pdfauthor={Niko Brummer},
              pdfstartpage=1, 
              pdfstartview=FitH,  
              pdfkeywords={}}

\makeatother

\usepackage[british]{babel}

\def\half{\frac{1}{2}}

\DeclareMathOperator{\trace}{tr}

\DeclareMathOperator{\vecf}{vec}

\def\expec#1{\left\langle#1\right\rangle}

\def\detm#1{\left\lvert#1\right\rvert}

\def\R{\mathbb{R}}
\def\mat#1{\begin{bmatrix}#1\end{bmatrix}}

\def\of#1{\!\left({#1}\right)}

\def\Xmat{\mathbf{X}}
\def\Smat{\mathbf{S}}
\def\Rmat{\mathbf{R}}
\def\Mmat{\mathbf{M}}
\def\Bmat{\mathbf{B}}
\def\Tmat{\mathbf{T}}
\def\Fmat{\mathbf{F}}

\def\xvec{\mathbf{x}}
\def\fvec{\mathbf{f}}

\def\muvec{\boldsymbol{\mu}}

\def\Lambdamat{\boldsymbol{\Lambda}}
\def\Thetamat{\boldsymbol{\Theta}}

\def\thetavec{\boldsymbol{\theta}}

\def\nulvec{\mathbf{0}}

\def\ND{\mathcal{N}}

\def\WD{\mathcal{W}}
\def\MW{\mathcal{MW}}

\def\model{\boldsymbol{\Phi}}

\begin{document}

\title{Generative, Fully Bayesian, Gaussian, Openset Pattern Classifier\\---Simplified Version---}
\def\today{BOSARIS Workshop---November 2012}

\author{Niko Br\"{u}mmer\\ AGNITIO Research, South Africa}

\maketitle

\noindent 

\section{Introduction}
Observable patterns live in $\R^N$ and belong to $K$ different classes. We are given a \emph{supervised training database}, $(\Xmat,L)$, where $\Xmat$ represents the training patterns of all the classes and $L$ represents their true class labels. We want to recognize the unknown classes to which new unlabelled \emph{test} patterns belong. For this purpose we pretend there is a generative model, with unknown parameter $\model$, that generates all train and test data. Although $\model$ is unknown, it is assumed to be the \emph{same} for train and test. 

For the purposes of this exercise, we choose perhaps the simplest possible such generative model, which has multivariate normal, class-conditional distributions of the form:
\begin{align}
P(\xvec|k,\model) &= \ND(\xvec|\muvec_k,\Lambdamat^{-1})
\end{align}
where $\xvec\in\R^N$ is a pattern, $k\in\{1,2,\ldots,K\}$ is a class index, $\muvec_k\in\R^N$ is the class conditional mean and $\Lambdamat^{-1}$ is the $N$-by-$N$, \emph{common within-class covariance} matrix. We shall refer to $\Lambdamat$ as the \emph{within-class precision}.  

The model parameters are collectively referred to as $\model$, where $\model=(\Mmat,\Lambdamat)$, and $\Mmat=\mat{\muvec_1&\muvec_2&\cdots&\muvec_K}$. 

Letting $\Pi$ denote some prior for $\model$, the fully Bayesian recipe\footnote{For a more complete derivation, see section 2.1 in Niko Brummer and Edward de Villiers, `Integrating out model parameters in generative and discriminative classifiers', 2011, available online at: \url{http://sites.google.com/site/nikobrummer/bayesian_model_integration.pdf}.} requires calculation of the \emph{parameter posterior} $P(\model|D,\Pi)$, which in turn gives the \emph{predictive distribution}:
\begin{align}
\label{eq:defpred}
P(\xvec|k,D,\Pi) = \int P(\xvec|k,\model) P(\model| D, \Pi) \,d\model
\end{align}
where $\xvec$ is a test pattern of hypothesized class $k$. This predictive distribution is the end goal of the exercise, since it can be used in a straight-forward calculation to find the classification posterior:
\begin{align}
\label{eq:class_post}
P(k|\xvec,D,\Pi,\pi) &= \frac{P_k P(\xvec|k,D,\Pi)}{\sum_{i=1}^K P_i P(\xvec|i,D,\Pi)}
\end{align}
where $\pi=(P_1,P_2,\ldots,P_K)$ is a given prior distribution over classes. Finally, the classification posterior can then be used to make minimum-expected-cost classification decisions.

In the rest of this document, we introduce notation for several probability distributions,  motivate the form of the parameter prior and then derive the parameter posterior and predictive distribution.

\section{Dramatis personae}
Here we introduce notation and properties of the probability distributions which will play the following roles in this problem:
\begin{quote}\begin{description}
  \item[likelihood:] product of multivariate Gaussians
  \item[prior for $\Lambdamat$:] Wishart
  \item[prior for $\Mmat$:] matrix normal, conditioned on $\Lambdamat$.
  \item[joint prior/posterior for $\Mmat,\Lambdamat$:] matrix normal Wishart
  \item[predictive distribution:] multivariate T.
\end{description}\end{quote}

\subsection{Multivariate Gaussian distribution}
\label{sec:gaussian}
The density of the multivariate Gaussian or normal distribution, for dimensionality $N$, defined in terms of mean $\muvec_k$ and precision $\Lambdamat$ is:
\begin{align}
\ND(\xvec|\muvec_k,\Lambdamat^{-1}) &= \frac{\detm{\Lambdamat}^{\half}}{(2\pi)^\frac{N}{2}}
\exp\left(-\frac{1}{2}(\xvec-\muvec_k)'\Lambdamat(\xvec-\muvec_k)\right)
\end{align}

Our likelihood will be expressed as a product of Gaussians. Let $\Xmat_k=\mat{\xvec_1&\xvec_2&\cdots&\xvec_{T_k}}$ represent $T_k$ iid samples from this density, then
\begin{align}
P(\Xmat_k|\muvec_k,\Lambdamat) &= (2\pi)^{-\frac{T_kN}{2}}\detm{\Lambdamat}^\frac{T_k}{2}
\exp(t_0+t_1+t_2)
\end{align}
where
\begin{align*}
t_0 &= -\half\trace(T_k\muvec_k\muvec_k'\Lambdamat), &
t_1 &= \trace(\fvec_k\muvec_k'\Lambdamat), &
t_2 &= -\half\trace(\Smat_k\Lambdamat)
\end{align*}
which we have expressed in terms of the first and second order stats:
\begin{align*}
\fvec_k &= \sum_{i=1}^{T_k} \xvec_i, & 
\Smat_k &= \Xmat_k\Xmat_k'
\end{align*}

\subsection{Wishart distribution}
\label{sec:Wishart}
\def\wish#1#2{\frac{\detm{\half#2}^\frac{#1}{2}}{\Gamma_N\of{\frac{#1}{2}}}}
We use the notation $\Lambdamat>0$ to indicate that $N$-by-$N$ matrix $\Lambdamat$ is positive definite. The probability density of the \emph{Wishart distribution} can be defined,\footnote{Our notation for the Wishart is parametrized for convenience in terms of $\Bmat$. In Bishop's book (appendix B), for example, the Wishart is defined in terms of $\Bmat^{-1}$.} for $\Lambdamat>0$, as:
\begin{align}
\WD(\Lambdamat|a,\Bmat) &= \frac{\detm{\half\Bmat}^\frac{a}{2}}{\Gamma_N\of{\frac{a}{2}}} \detm{\Lambdamat}^\frac{a-N-1}{2} \exp\left(-\half\trace(\Bmat\Lambdamat)\right) 
\end{align}
where $a>N-1$; $\Bmat$ is $N$-by-$N$ positive definite; and $\Gamma_N$ is the \emph{multivariate gamma function}, defined as:
\begin{align}
\label{eq:mgamma}
\Gamma_N(x) &= \pi^\frac{N(N-1)}{4} \prod_{i=1}^N \Gamma\of{x+\frac{1-i}{2}} 
\end{align}
with $\Gamma(x)=\Gamma_1(x)$ the usual gamma function. The expected value of the Wishart density is $\expec{\Lambdamat}= a\Bmat^{-1}$.

Since the Wishart density integrates to one, we find the useful result:
\begin{align}
\label{eq:wishart_integral}
\int_{\Lambdamat>0} \detm{\Lambdamat}^\frac{a-N-1}{2} \exp\left(-\half\trace(\Bmat\Lambdamat)\right) \,d\Lambdamat = \Gamma_N\of{\frac{a}{2}}\detm{\half\Bmat}^{-\frac{a}{2}}
\end{align}

The Wishart density will form the prior for the within-class precision, $\Lambdamat$.

 \subsection{Matrix Normal}
\def\MD{\mathcal{M}}
\def\Imat{\mathbf{I}}
The prior for the class means, $\Mmat=(\muvec_1,\ldots,\muvec_K)$, will be formed by a matrix normal density. We could instead use a product of independent normal distributions for each $\mu_k$, but the matrix normal conveniently models the whole matrix.

The \emph{matrix normal} density, for an $N$-by-$K$ matrix $\Mmat$, can be expressed as:
\begin{align}
\MD(\Mmat|\Thetamat,\Rmat,\Lambdamat)
&= \frac{\detm{\Rmat}^\frac{N}{2}\detm{\Lambdamat}^\frac{K}{2}}{(2\pi)^\frac{NK}{2}}
\exp\of{-\half\trace\bigl(\Rmat(\Mmat-\Thetamat)'\Lambdamat(\Mmat-\Thetamat)\bigr)}
\end{align}
where the location parameter, $\Thetamat$ is $N$-by-$K$ and where there are two positive definite precision parameters: $\Rmat$, $K$-by-$K$ and $\Lambdamat$, $N$-by-$N$. The matrix normal is related to the multivariate Gaussian as follows:\footnote{Here $\vecf$ stacks the columns of a matrix into a single vector and $\otimes$ denotes the Kronecker matrix product. Keep in mind that $(\Rmat\otimes\Lambdamat)^{-1}=\Rmat^{-1}\otimes\Lambdamat^{-1}$.}
\begin{align}
\MD(\Mmat|\Thetamat,\Rmat,\Lambdamat) &= \ND(\vecf(\Mmat)|\vecf(\Thetamat),\Rmat^{-1}\otimes\Lambdamat^{-1})
\end{align}

For our prior, we won't use the full power of the matrix normal. We set $\Thetamat=\nulvec$ and $\Rmat=r\Imat$.

\subsubsection{Marginal}
\label{sec:mnmarg}
Let $c_k$ denote the $k$-th element on the diagonal of $\Rmat^{-1}$. Then we can express the marginal for column $k$ of $\Mmat$ as:\footnote{See e.g. Bishop's book, equation (B.51).}
\begin{align}
P(\muvec_k|\Thetamat,\Rmat,\Lambdamat) = \ND(\muvec_k|\thetavec_k,c_k\Lambdamat^{-1})
\end{align}
where $\thetavec_k$ is column $k$ of $\Thetamat$. For the above prior parameters we have $\thetavec_k=\nulvec$ and $c_k=\frac{1}{r}$.

\subsection{Matrix Normal Wishart}
\label{sec:mnw}
For $\Mmat$, $N$-by-$K$ and $\Lambdamat$, $N$-by-$N$ positive definite, the joint \emph{matrix normal Wishart} density can be expressed as:
\begin{align}
\label{eq:mwform}
\begin{split}
&\MW(\Mmat,\Lambdamat|\Thetamat,\Rmat,a,\Bmat) 
=\MD(\Mmat|\Thetamat,\Rmat,\Lambdamat)\WD(\Lambdamat|a,\Bmat) \\
&= \frac{\detm{\Rmat}^\frac{N}{2}\detm{\Lambdamat}^\frac{K}{2}}{(2\pi)^\frac{NK}{2}}
\exp\of{-\half\trace\bigl(\Rmat(\Mmat-\Thetamat)'\Lambdamat(\Mmat-\Thetamat)\bigr)}
\WD(\Lambdamat|a,\Bmat) \\
&= \wish{a}{\Bmat} \frac{\detm{\Rmat}^\frac{N}{2}}{(2\pi)^\frac{NK}{2}}
\detm{\Lambdamat}^\frac{a+K-N-1}{2}\exp(e_1+e_2+e_3)
\end{split}
\end{align}
where
\begin{align*}
e_1 &= -\half\trace(\Rmat\Mmat'\Lambdamat\Mmat), &
e_2 &= -\half\trace\bigl((\Bmat+\Thetamat\Rmat\Thetamat')\Lambdamat\bigr),&
e_3 &= \trace(\Thetamat\Rmat\Mmat'\Lambdamat)
\end{align*}
and where the parameters $\Thetamat,\Rmat,a,\Bmat$ are as introduced above.

\subsection{Multivariate T}
\def\dvec{\mathbf{d}}
\def\TD{\mathcal{T}}
Below, when expressing the predictive distribution, we shall need the solution to an integral of the form:
\begin{align}
\begin{split}
&\int_{\Lambdamat>0} \ND(\xvec|\thetavec,\beta^{-1}\Lambdamat^{-1}) \WD(\Lambdamat|a,\Bmat)
\,d\Lambdamat \\
&= \frac{\detm{\half\Bmat}^\frac{a}{2} \beta^\frac{N}{2}}
{(2\pi)^\frac{N}{2}\Gamma_N\of{\frac{a}{2}}}
\int \detm{\Lambdamat}^\frac{a+1-N-1}{2} 
\exp\of{-\half\trace\bigl((\beta\dvec\dvec'+\Bmat)\Lambdamat\bigr)}
\,d\Lambdamat
\end{split}
\end{align}
where we have defined $\dvec=\xvec-\thetavec$ for convenience. We solve the integral using ~\eqref{eq:wishart_integral}, and then simplify using~\eqref{eq:mgamma} and the matrix determinant lemma:\footnote{$\detm{\Bmat+\beta\dvec\dvec'}=(1+\beta\dvec'\Bmat^{-1}\dvec)\detm{\Bmat}$}
\begin{align}
\label{eq:Tderivation}
\begin{split}
&\int_{\Lambdamat>0} \ND(\xvec|\thetavec,\beta^{-1}\Lambdamat^{-1}) \WD(\Lambdamat|a,\Bmat)
\,d\Lambdamat \\
&= \frac{\detm{\half\Bmat}^\frac{a}{2} \beta^\frac{N}{2}}
{(2\pi)^\frac{N}{2}\Gamma_N\of{\frac{a}{2}}}
\Gamma_N\of{\frac{a+1}{2}}\detm{\half(\beta\dvec\dvec'+\Bmat)}^{-\frac{a+1}{2}} \\
&= \left(\frac{\beta}{\pi}\right)^\frac{N}{2} 
\frac{\Gamma_N\of{\frac{a+1}{2}}}{\Gamma_N\of{\frac{a}{2}}}
\frac{\detm{\Bmat}^\frac{a}{2}}{\detm{\beta\dvec\dvec'+\Bmat}^\frac{a+1}{2}} \\
&= \left(\frac{\beta}{\pi}\right)^\frac{N}{2} 
\frac{\Gamma\of{\frac{a+1}{2}}}{\Gamma\of{\frac{a+1-N}{2}}}
\frac{\detm{\Bmat}^\frac{a}{2}}{\detm{\beta\dvec\dvec'+\Bmat}^\frac{a+1}{2}} \\
&= \frac{\Gamma\of{\frac{a+1}{2}}}{\Gamma\of{\frac{a+1-N}{2}}} 
\detm{\frac{\pi}{\beta}\Bmat}^{-\half}
\Bigl(1+\beta\dvec'\Bmat^{-1}\dvec\Bigr)^{-\frac{a+1}{2}} \\
&= \TD_N(\xvec|\thetavec,\beta^{-1}\Bmat,a)
\end{split}
\end{align}
This is a \emph{multivariate T distribution}, for which we have introduced the notation\footnote{Again, our notation is for convenience here. It differs (just cosmetically) from the way Bishop (his appendix B), defines his `multivariate Student's t' and also (again cosmetically) from the `Box and Tiao T distribution' in Minka's `Inferring a Gaussian'.} $\TD_N$.

\section{Parameter inference}
Given the supervised database $(\Xmat,L)$ and a conjugate prior, we do a Bayesian inference of the parameters $\model=(\Mmat,\Lambdamat)$. The likelihood is the product of Gaussians of all the data in $\Xmat$. We use the matrix normal Wishart as conjugate prior and obtain a posterior of the same form.

\subsection{Likelihood}
For the purpose of parameter inference, the supervised database is represented by the triple statistic of the form $T_k, \fvec_k, \Smat_k$, for each class $k\in\{1,2,\ldots,K\}$. For convenience we define: $\Tmat$ to be the diagonal matrix, with diagonal elements $T_1,T_2,\ldots,T_K$; $T=\trace(\Tmat)$ is the total number of patterns; $\Smat=\sum_k \Smat_k$ and $\Fmat=\mat{\fvec_1\cdots\fvec_K}$. Recall that the mean parameters are represented as $\Mmat=\mat{\muvec_1\cdots\muvec_K}$.

Recalling section~\ref{sec:gaussian}), the parameter likelihood is:
\begin{align}
\label{eq:lh4post}
L(\Mmat,\Lambdamat|D) &= \detm{\Lambdamat}^\frac{T}{2}
\exp(E_1+E_2+E_3)
\end{align}
where
\begin{align*}
E_1 &= -\half\trace(\Tmat\Mmat'\Lambdamat\Mmat), &
E_2 &= -\half\trace(\Smat\Lambdamat), &
E_3 &= \trace(\Fmat\Mmat'\Lambdamat)
\end{align*}

\subsection{Parameter prior}
\label{sec:param_prior}
We assign a matrix normal Wishart prior, with a zero location parameter. Letting $\Pi=(\Rmat,a,\Bmat)$, our conjugate prior is of the form (recall section~\ref{sec:mnw}):
\begin{align}
\begin{split}
P(\Mmat,\Lambdamat|\Pi) &= \MW(\Mmat,\Lambdamat|\mathbf{0},\Rmat,a,\Bmat) \\
&\propto \detm{\Lambdamat}^\frac{a+K-N-1}{2}\exp(e_1+e_2)
\end{split}
\end{align}
where
\begin{align*}
e_1 &= -\half\trace(\Rmat\Mmat'\Lambdamat\Mmat), &
e_2 &= -\half\trace(\Bmat\Lambdamat)
\end{align*}
We choose $\Rmat=r\Imat$, where larger $r$ forces the means closer to the origin.

\subsection{Parameter posterior}
\label{sec:param_post}
The parameter posterior can now be derived as:
\begin{align}
\label{eq:postform}
\begin{split}
P(\Mmat,\Lambdamat|D,\Pi) 
&\propto P(\Mmat,\Lambdamat|\Pi) L(\Mmat,\Lambdamat|D)  \\
&\propto \detm{\Lambdamat}^\frac{a+T+K-N-1}{2} 
\exp(s_1+s_2+s_3)
\end{split}
\end{align}
where
\begin{align*}
s_1 &= -\half\trace\bigl((\Rmat+\Tmat)\Mmat'\Lambdamat\Mmat\bigr), &
s_2 &= -\half\trace\bigl((\Smat+\Bmat)\Lambdamat\bigr), &
s_3 &= \trace(\Fmat\Mmat'\Lambda) 
\end{align*}
which also has a matrix normal Wishart form, so that:
\begin{align}
P(\Mmat,\Lambdamat|D,\Pi) 
&= \MW(\Mmat,\Lambdamat|\Mmat^*,\Rmat^*,a^*,\Bmat^*)
\end{align}
where we can identify the parameters by comparing~\eqref{eq:postform} with~\eqref{eq:mwform}:
\begin{align*}
\Mmat^* &= \Fmat (\Rmat^*)^{-1}, &
\Rmat^* &= r\Imat + \Tmat\\
a^* &= a+T, &
\Bmat^* &= \Bmat + \Smat - \Fmat(\Rmat^*)^{-1}\Fmat' 
\end{align*}
Notice that if $r=0$, the class means are just the data averages. At the other extreme, $r\to\infty$, the class means remain at zero.

\section{Predictive distribution}
We can now finally derive an expression for the predictive distribution, $P(\xvec|k,\Xmat,L,\Pi)$, as defined in~\eqref{eq:defpred}. Here $k$ is the hypothesized class of a new test pattern $\xvec$; $(\Xmat,L)$ is the supervised training database and $\Pi$ represents all prior assumptions and parameters. As mentioned in the introduction, the predictive distribution is end goal of the whole exercise. 

Thanks to the conjugacy, the integral over the parameter posterior can be found in closed form: 
\begin{align}
\begin{split}
&P(\xvec|k,\Xmat,L,\Pi) \\
&= \int P(\xvec|k,\model) P(\model| D, \Pi) \,d\model \\
&= \int \ND(\xvec|\muvec_k,\Lambdamat^{-1}) \MW(\Mmat,\Lambdamat|\Mmat^*,\Rmat^*,a^*,\Bmat^*) \,d\muvec_1 \cdots d\muvec_K\,d\Lambdamat \\
&= \int \ND(\xvec|\muvec_k,\Lambdamat^{-1}) \ND(\muvec_k|\muvec^*_k,c^*_k\Lambdamat^{-1}) \,d\muvec_k \WD(\Lambdamat|a^*,\Bmat^*) \,d\Lambdamat
\end{split}
\end{align}
where we used the result of section~\ref{sec:mnmarg}, with $c^*_k$ the $k$-th diagonal element of $(\Rmat^*)^{-1}$ and $\muvec^*_k$ the $k$-th column of $\Mmat^*$. Next, we integrate out $\muvec_k$ by simply adding variances and finally use~\eqref{eq:Tderivation} to integrate out $\Lambdamat$:
\begin{align}
\label{eq:pd}
\begin{split}
P(\xvec|k,D,\Pi) &= \int \ND(\xvec|\muvec^*_k,(c^*_k+1)\Lambdamat^{-1}) \WD(\Lambdamat|a^*,\Bmat^*) \,d\Lambdamat \\
&= \TD_N(\xvec|\muvec^*_k,(c^*_k+1)\Bmat^*,a^*)
\end{split}
\end{align}

\subsection{At non-informative prior}
We let $a\to0$, so that $a^* = T$ and $\Bmat\to\nulvec$, so that $\Bmat^* = \Smat - \Fmat(\Rmat^*)^{-1}\Fmat'$. For recognizing the class of $\xvec$, via the posterior~\eqref{eq:class_post}, we can ignore all factors of~\eqref{eq:pd} which are not conditioned on the class. This gives:
\begin{align}
P(\xvec|k,D,\Pi) &\propto (c^*_k+1)^{-\frac{N}{2}}\Biggl(1+
\frac{(\xvec-\muvec^*_k)'(\Bmat^*)^{-1}(\xvec-\muvec^*_k)}{c^*_k+1}
\Biggr)^{-\frac{T+1}{2}} 
\end{align}
where $\muvec^*_k=\frac{1}{r+T_k}\fvec_k$ and $c^*_k=\frac{1}{r+T_k}$.

Notice that as long as $r>0$, we can also get a predictive distribution for a class with no training data (with $T_k=0$). Adding such a class will leave $\Bmat^*$ and $a^*$ unchanged, with  $c^*_k=\frac1r$ and $\muvec_k=\nulvec$.

\section{Model evidence}
Here we assume $a,\Bmat$ are given (we will eventually take them to the non-informative limits at zero), but we are interested in how the model changes as a function of $r$, so that we can decide which value to use for it, possibly using an ML or MAP estimate. Recall that the supervised database is denoted $\Xmat,L$, where $\Xmat$ represents the data for all the classes and $L$ represents the class labels. We need to compute the model \emph{evidence}:
\begin{align}
\begin{split}
P(\Xmat|r,L,a,\Bmat) &= \int P(\Xmat,\Mmat,\Lambdamat|r,L,a,\Bmat) \,d\Mmat\,d\Lambdamat \\
&= \int P(\Xmat|\Mmat,\Lambdamat,L) P(\Mmat,\Lambdamat|r,a,\Bmat)\,d\Mmat\,d\Lambdamat 
\end{split}
\end{align}
Here $P(\Mmat,\Lambdamat|r,a,\Bmat)$ is the matrix normal Wishart parameter prior as defined in section~\ref{sec:mnw}. Omitting the constant $(2\pi)^{-\frac{NK}{2}}$, we have:
\begin{align}
\begin{split}
P(\Mmat,\Lambdamat|r,a,\Bmat) &\propto 
\wish{a}{\Bmat} \\
&\;\;\;\detm{\Rmat}^\frac{N}{2}
\detm{\Lambdamat}^\frac{a+K-N-1}{2}
\exp\Bigl[
-\frac12\trace(\Rmat\Mmat'\Lambdamat\Mmat)
-\frac12\trace(\Bmat\Lambdamat)
\Bigr]
\end{split}
\end{align}
Recall that $\Rmat=r\Imat$. Since we are interested only in inferring $r$, we could omit $\wish{a}{\Bmat}$, but we show this factor here to stress the fact that if $a=0$ and $\Bmat=\nulvec$, then the prior and hence also the evidence becomes improper. 

The other factor in the integrand is the likelihood, given by~\eqref{eq:lh4post}:
\begin{align}
P(\Xmat|\Mmat,\Lambdamat,L) &\propto
\detm{\Lambdamat}^\frac{T}{2}\exp\Bigl[
-\half\trace(\Tmat\Mmat'\Lambdamat\Mmat)
-\half\trace(\Smat\Lambdamat)
+\trace(\Fmat\Mmat'\Lambdamat)
\Bigr]
\end{align}
Multiplying prior and likelihood, we get the integrand:
\begin{align}
\begin{split}
\mathcal{I}(\Mmat,\Lambdamat) &= \detm{\Rmat}^\frac{N}{2}
\detm{\Lambdamat}^\frac{T+a+K-N-1}{2} \\
&\;\;\;\;\exp\Bigl[
-\half\trace((\Tmat+\Rmat)\Mmat'\Lambdamat\Mmat)
-\half\trace((\Smat+\Bmat)\Lambdamat)
+\trace(\Fmat\Mmat'\Lambdamat)
\Bigr]
\end{split}
\end{align}
This is proportional to the matrix normal Wishart parameter posterior that we found above, with parameters $a^*,\Bmat^*,\Mmat^*,\Rmat^*$. If we nomalize this, the integral is one, so that\footnote{omitting factors independent of $r$}:
\begin{align}
\begin{split}
\int\mathcal{I}(\Mmat,\Lambdamat) \,d\Mmat \,d\Lambdamat
&\propto\detm{\Rmat}^\frac{N}{2}
\detm{\frac12\Bmat^*}^{-\frac{a^*}{2}}
\detm{\Rmat^*}^{-^\frac{N}{2}}
\end{split}
\end{align}
We can simplify:
\begin{align}
\frac{\detm{\Rmat}}{\detm{\Rmat^*} }
&= \frac{r^K}{\prod_{k=1}^K(r+T_i)}
\end{align} 

\subsection{At non-informative prior}
At $a=0,\Bmat=\nulvec$, the log evidence is:
\begin{align}
\frac{N}{2}\Bigl[K\log(r)-\sum_{i=1}^K\log(r+T_k)\Bigr]
-\frac{T}{2}\log\detm{\Smat-\sum_{i=1}^K\fvec_k\fvec_k'\frac{1}{r+T_k}}
\end{align}

\end{document}